\documentclass{article}
\usepackage{ijcai19}
\usepackage{titlesec}
\usepackage{amsfonts}
\usepackage{times}
\usepackage{soul}
\usepackage{url}
\usepackage{hyperref}
\usepackage[small]{caption}
\usepackage{graphicx}
\usepackage{multirow}
\usepackage{multicol}
\usepackage{bigstrut}
\usepackage{amsmath}
\usepackage{amsthm}
\usepackage{booktabs}
\usepackage{algorithm}
\usepackage{algorithmic}
\usepackage{subfigure}
\usepackage{bm}
\usepackage{setspace}
\newtheorem{definition}{Definition}
\title{Enhancing Item Response Theory for Cognitive Diagnosis}

\author{
	Song Cheng$^1$
	\and
	Qi Liu$^1$
	\affiliations
	$^1$Department of Computer Science, University of Science and Technology of China, China\\
	\emails
	chsong@mail.ustc.edu.cn,
	qiliuql@ustc.edu.cn
}

\begin{document}
	\maketitle
	
	\begin{abstract}
		Cognitive diagnosis is a fundamental and crucial task in many educational applications, e.g., computer adaptive test and cognitive assignment. {\em Item Response Theory} (IRT) is a classical cognitive diagnosis method which can provide interpretable parameters (i.e., student latent trait, question discrimination and difficulty) for analyzing student performance. However, traditional IRT ignores the rich information in question texts, cannot diagnose knowledge concept proficiency, and it is inaccurate to diagnose the parameters for the questions which only appear several times. To this end, in this paper, we propose a general {\em Deep Item Response Theory} (DIRT) framework to enhance traditional IRT for cognitive diagnosis by exploiting semantic representation from question texts with deep learning. In DIRT, we first use a proficiency vector to represent students' proficiency on knowledge concepts, and embed question texts and knowledge concepts to dense vectors by {\em Word2Vec}. Then, we design a deep diagnosis module to diagnose parameters in traditional IRT by deep learning techniques. Finally, with the diagnosed parameters, we input them into logistic-like formula of IRT to predict student performance. Extensive experimental results on real-world data clearly demonstrate the effectiveness and interpretation power of DIRT framework.
	\end{abstract}
\section{Introduction}
Many education systems such as intelligent tutoring systems and massive open online course, provide a series of computer aided applications, e.g., computer adaptive test~\cite{gershon2005computer} and cognitive assessment~\cite{tatsuoka2009cognitive}.  Among these applications, cognitive diagnosis which aims at discovering the latent traits or characteristics of students is a fundamental and crucial task. For instance, diagnosing student knowledge concept proficiency~\cite{wu2015cognitive,Wu2017KnowledgeOG}, knowledge tracing~\cite{Chen2017TrackingKP}.
\begin{figure}
	\setlength{\abovecaptionskip}{0.2cm}
	\setlength{\belowcaptionskip}{-0.53cm} 
	\centering
	\includegraphics[width=3in,height=1.5in]{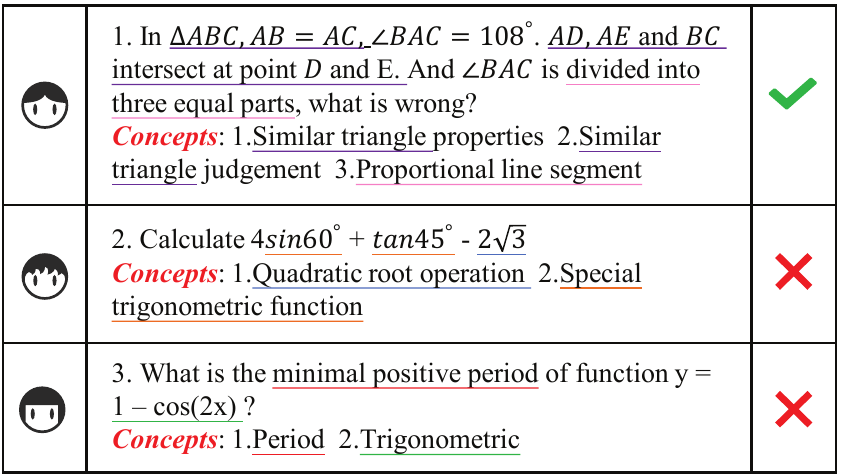}
	\caption{A toy example of student question records}
	\label{fig:examples}
\end{figure}

The well-known {\em Item Response Theory} (IRT)~\cite{embretson2000item} which roots in psychological measurement theory is a classical cognitive diagnosis method. It can predict student performance on the questions by the logistic-like IRT formula and provide interpretable parameters (i.e., student latent trait, discrimination and difficulty of questions) for performance analyzing. Since the strong interpretability of these parameters, IRT has been widely applied in many applications~\cite{hambleton1989principles}, such as personalized adaptive test~\cite{van2000computerized}, cognitive assessment~\cite{tatsuoka2009cognitive}.

Though IRT has made great success in cognitive diagnosis,  there are still three issues limiting its usefulness. First, traditional IRT can only exploit student response results, and ignores the rich information and the associations between question texts and knowledge concepts. As shown in Figure~\ref{fig:examples}, the question texts and the knowledge concepts on the underline with the same color are closely related, which is helpful for modeling questions~\cite{huang2017question}. Thus it is critical to improve IRT to exploit semantic representation from question texts. Second, IRT cannot diagnose parameters accurately for the questions which only have several records. Since students often meet these rare questions in real scene, therefore, improving IRT to have strong robustness for rare questions is important. Third, IRT cannot diagnose student proficiency on knowledge concepts, it only provides a overall latent trait for students, while each question usually assesses different knowledge concepts. Although its extended model {\em Multidimensional Item Response Theory} (MIRT)~\cite{yao2006multidimensional} can diagnose student proficiency on knowledge concepts, it is sensitive to the concepts on which students have high proficiency~\cite{yao2006multidimensional}. Thus, enhancing IRT to provide diagnosis results on each knowledge concept in a reliable way is still an open problem.

To this end, in this paper, we propose a general {\em deep item response theory} (DIRT) framework to enhance IRT for cognitive diagnosis by deep learning. It is composed of three modules: input, deep diagnosis and prediction module. Specifically, in input module, for diagnosing each knowledge concept proficiency for students, we use a proficiency vector to represent the student proficiency on each knowledge concept, and embed question texts and knowledge concepts to dense vectors by {\em World2Vec}~\cite{mikolov2013distributed}. In deep diagnosis module, since deep learning has great feature representation ability, and can automatically represent features in the same domain robustly~\cite{kim2013deep,zhang2016semantics}, we use deep learning methods to diagnose student latent trait, discrimination and difficulty of questions by exploiting semantic representation from question texts, and enhance the robustness for rare questions.
In prediction module, we input the parameters diagnosed by deep diagnosis module into the logistic-like IRT formula to predict student performance. Extensive experimental results on real-world data clearly demonstrate the effectiveness and interpretation power of DIRT framework.


\section{Preliminary}
In this section, we first give the definition of cognitive diagnosis in this paper. Then we briefly review the classical cognitive diagnosis method IRT and its extended method MIRT.
\subsection{Problem Definition}
Suppose there are $L$ students, $M$ questions and total $P$ knowledge concepts. Student question records are represented by $R=\{R_{ij}\arrowvert 1\le i\le L, 1\le j\le M\}$, where $R_{ij}=\langle S_i, Q_j, r_{ij}\rangle$ denotes the student $S_i$ gets score $r_{ij}$ on question $Q_j$. $Q_j=\langle QT_j, QK_j \rangle$ is composed of question texts $QT_j$ and corresponding knowledge concepts $QK_j$. 
\begin{definition}
	\textbf{(Cognitive Diagnosis)}. Given the question records $R$ of student $S$, our goal is to build a uniform model $\mathcal{M}$ to diagnose the proficiency vector $\bm{\alpha}$ on all knowledge concepts $QK$ for each student. 
\end{definition}
Since there is no ground truth for diagnosis results, and generally, the more accurate the model $\mathcal{M}$ predicts on student performance prediction task, the better the diagnosis results are~\cite{wu2015cognitive}, so we use performance prediction task to validate the effectiveness of cognitive diagnosis results.
\subsection{Item Response Theory}
\subsubsection{Definition}
IRT is one of the most important psychological and educational theories and it roots in psychological measurement~\cite{embretson2000item}. With the student latent trait $\theta$, discrimination $a$ and difficulty $b$ of the question, IRT can predict the probability that the student answers the question correctly by logistic-like IRT formula which is defined as follow:
\begin{equation}
\setlength{\abovedisplayskip}{1pt}
\setlength{\belowdisplayskip}{1pt}
P(\theta)=\frac{1}{1 + e^{-Da(\theta -b)}},
\label{eq:IRT}
\end{equation}
where $P(\theta)$ is the correct probability, $D$ is a constant which often set as $1.7$. The more details of latent trait $\theta$, discrimination $a$ and difficulty $b$ are as follows:
\paragraph{Latent Trait.} Latent trait $\theta$ of a student represents student's overall ability and has the same value for all questions. There is no theoretical limitation on the range of latent trait.
\paragraph{Discrimination.} Discrimination $a$ is the ability of the questions to divide students into different levels. The theoretical range of $a$ is $[-4,4]$.
\paragraph{Difficulty.} Difficulty $b$ means how difficult the question is, and the theoretical range of $b$ is also $[-4,4]$.
\subsubsection{Shortcomings of Item Response Theory}
The latent trait $\theta$, discrimination $a$ and difficulty $b$ described above are usually estimated by probability algorithms such as {\em Maximum Likelihood Estimation} (MLE) or {\em Expectation Maximization} (EM) only with students' responses $r$ in students' question records $R$. Thus, IRT cannot capture the rich semantic information in question texts $QT$. In addition, IRT is not robust enough to diagnose parameters $a$ and $b$ accurately for questions which only have several records, and IRT only diagnose student latent trait on question level, while cannot provide more detailed diagnosis results on concepts.
\subsection{Multidimensional Item Response Theory}
MIRT is extended from IRT, and its purpose is to meet the demands of multidimensional data~\cite{yao2006multidimensional}. With the student multidimensional latent traits $\bm{\theta}=(\theta_1,\theta_2,...,\theta_m)^T$ on each knowledge concepts, the knowledge concept discriminations $\bm{a}=(a_1,a_2,...,a_m)^T$ and the intercept term $d$ of the question, MIRT can also predict the probability of the student answers the question correctly by MIRT formula which is defined as follow:
\vspace{-0.1em}
\begin{equation}
\vspace{-0.1em}
\begin{aligned}
\bm{P(\theta)}=\frac{e^{\bm{a}^T\bm{\theta} +d}}{1+e^{\bm{a}^T\bm{\theta} +d}},
\end{aligned}
\label{eq:MIRT}
\end{equation}
where $P(\theta)$ is the correct probability the same as IRT.

Since the process of estimating parameters for MIRT is same as IRT, therefore, MIRT has the same shortcomings as IRT. Although MIRT can provide knowledge concept proficiency for students, it is sensitive to the knowledge concepts on which students have high latent trait~\cite{yao2006multidimensional}.

\begin{figure}
	\setlength{\abovecaptionskip}{0.2cm}
	\setlength{\belowcaptionskip}{-0.5cm}
	\centering
	\includegraphics[height=1.4in]{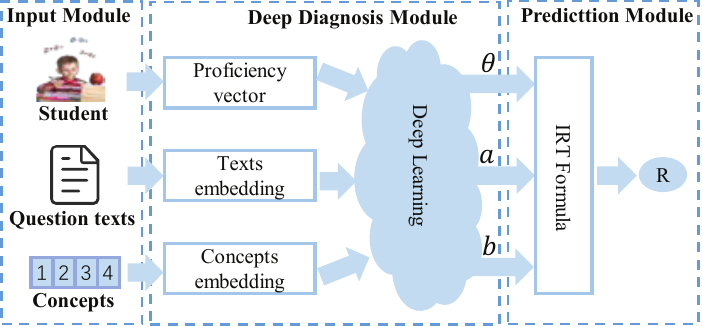}
	\caption{General DIRT Framework.}
	\label{fig:general-framework}
\end{figure}
\section{DIRT Framework}
To overcome the shortcomings in IRT and MIRT, we propose a general DIRT framework which shown in Figure~\ref{fig:general-framework} by enhancing IRT with deep learning to obtain latent trait, discrimination and difficulty in traditional IRT. DIRT contains three modules, i.e., input module, deep diagnosis module and prediction module. As shown in Figure~\ref{fig:general-framework}, the input module initializes a proficiency vector which can provide diagnosis results on each knowledge concept for each student, and embeds question texts and knowledge concepts to vectors. In deep diagnosis module, latent trait, discrimination and difficulty are obtained by exploiting question texts with deep learning techniques(e.g., DNN and LSTM) to enhance model robustness. The prediction module predicts the probability the student answers the question correctly by the logistic-like IRT formula with parameters diagnosed by deep diagnosis module. We give a specific implementation in the section bellow.

\subsection{A Specific Implementation of DIRT}
In this section, we give a specific implementation of DIRT framework, we first introduce the details of this implementation, and then we specify the learning strategy of it. 
\subsubsection{Input Module} 
Given a student $S$, in order to diagnose her proficiency on all $P$ knowledge concepts. We use a proficiency vector $\bm{\alpha}=(\alpha_1,\alpha_2,...,\alpha_P)$ to represent the proficiency on each knowledge concept for the student, where $\alpha_l\in [0,1]$ represents the degree a student masters the knowledge concept $l$. The proficiency vector $\bm{\alpha}$ is initialized randomly by normal distribution.

For the question $Q=\langle QT, QK \rangle$, the question texts are described with a sequence of words $QT=\{w_1,w_2,...,w_u\}$, where $u$ is the length of $QT$, and $w_i \in \mathbb{R}^{d_0}$ is initialized with a $d_0$-dimensional pre-trained word embedding with {\em Word2Vec}~\cite{mikolov2013distributed} method. The knowledge concepts are represented by onehot vectors $QK=\{K_1,K_2,...,K_v\}$, $K_i\in \{0,1\}^{P}$, where $v$ is the number of knowledge concepts associate to the question $Q$. Since onehot representation is very sparse for neural network training, we utilize a $d_1$-dimension dense layer to acquire the dense embedding~\cite{guo2017deepfm} for each knowledge concept $K_i$. We sign the dense embedding of $K_i$ as $k_i$:
\begin{equation}
\setlength{\abovedisplayskip}{1pt}
\setlength{\belowdisplayskip}{1pt}
k_i = K_i\textbf{W}_k,
\label{eq:denseweight}
\end{equation}
where $\textbf{W}_k\in\mathbb{R}^{P\times d_1}$ are the parameters of the $d_1$-dimension dense layer, and $k_i \in\mathbb{R}^{d_1}$.
\subsubsection{Deep Diagnosis Module}
\begin{figure}
	\centering
	\includegraphics[scale=0.8]{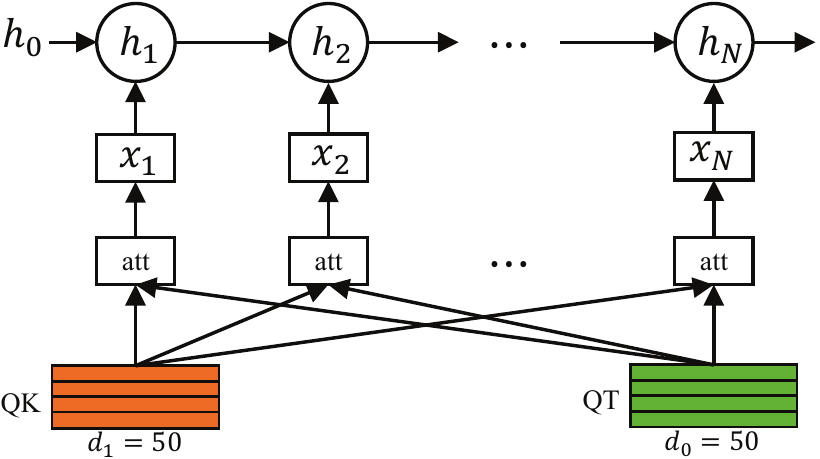}
	\caption{Attention-based LSTM in DIRT Framework.}
	\label{fig:att-lstm}
	\vspace{-1em}
\end{figure}

Deep diagnosis module is mainly achieved by deep learning techniques (e.g., DNN, LSTM), that is because deep learning has the great texts auto representation ability, and can exploit texts information from semantic perspective and enhance the robustness of the framework. In this module, we diagnose latent trait, discrimination and difficulty which have strong interpretability in traditional IRT. The details are as follows.
\paragraph{Latent Trait.} Latent trait in IRT has strong interpretability for students' performance on questions. We diagnose latent trait parameter in this part. The latent trait $\theta$ is closely related to the proficiency of knowledge concepts~\cite{yao2006multidimensional}, students who have great proficiency on knowledge concepts will have high latent trait because concepts proficiency can reflect the comprehensive ability of students. Because of the great feature representation ability of deep learning techniques, we use a deep neural network (DNN) to diagnose latent trait $\theta$. Specifically, given the proficiency vector $\bm{\alpha}=(\alpha_1,\alpha_2,...,\alpha_P)$ of the student $S$ and a question $Q$, we multiply the corresponding proficiency in $\bm{\alpha}$ with the concepts dense embedding of the question, by doing this, we will get a $d_1$-dimension vector $\Theta\in\mathbb{R}^{ d_1}$, and then we input $\Theta$ into DNN to model the latent trait automatically, which is defined as follow:
\begin{equation}
\setlength{\abovedisplayskip}{1pt}
\setlength{\belowdisplayskip}{1pt}
\theta = \text{DNN}_{\theta}(\Theta),\quad \Theta = \sum_{k_i \in \mathcal{K}_q}\alpha_i k_i,
\label{eq:latenttrait}
\end{equation}
where $\mathcal{K}_q$ is the set of the dense embedding  of $v$ knowledge concepts. Latent trait diagnosed here is different for each question, which has stronger interpretability than it in IRT.
\paragraph{Discrimination.} Discrimination $a$ can be applied to analyze students' performance distribution on the question. As for the method to obtain discrimination parameter, inspired by the relationship between Multidimensional Item Discrimination (MDISC) and knowledge concepts~\cite{yao2006multidimensional}, we learn question discrimination $a$ from knowledge concepts corresponded to the question. Here, we use another DNN to diagnose question discrimination $a$ with the same reasons as latent trait diagnosing. Specifically, we sum the dense embedding of knowledge concepts in $\mathcal{K}_q$ to get a $d_1$-dimensional vector $A\in\mathbb{R}^{d_1}$, and then, we input $A$ into the DNN to diagnosis question discrimination. Because the theoretical range of $a$ is $[-4,4]$, we use the $sigmoid$ function to minus $0.5$ and multiply $8$ to meet the range requirement. The definition of $a$ is as follow:
\begin{equation}
\setlength{\abovedisplayskip}{1pt}
\setlength{\belowdisplayskip}{1pt}
a = 8\times\text{sigmoid}(\text{DNN}_{a}(A) - 0.5),\quad A = \sum_{k_i \in \mathcal{K}_q}k_i,
\label{eq:discrimination}
\end{equation}
where the structure of $\text{DNN}_a$ is same as $\text{DNN}_{\theta}$, but the parameters are not shared between them.
\begin{figure}
	\centering
	\includegraphics[scale=0.9]{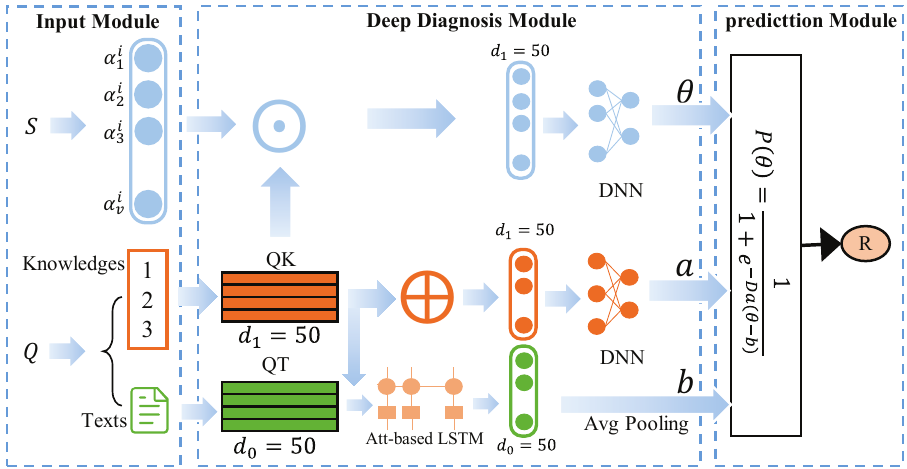}
	\caption{The Structure of the Specific Implementation of DIRT.}
	\label{fig:model}
	\vspace{-1em}
\end{figure}
\paragraph{Difficulty.} Difficulty $b$ determines how difficult the question is. The first perspective to diagnosis question difficulty is exploiting question texts, because the difficulty is closely related to question texts~\cite{huang2017question}, we can model question difficulty is by exploiting question texts. In this term, LSTM can handle and represent long time sequence texts from semantic perspective which can provide strong robustness for rare questions, so we use LSTM to model difficulty $b$ from text perspective. As for the second perspective, the depth and width of knowledge concepts examined by the question also have a great impact on difficulty. The deeper and wider the knowledge concepts are examined, the more difficult the question is. Obviously, the depth and width of the examined concepts can be reflected by the relevance between question texts and the corresponding concepts. As shown in Figure~\ref{fig:examples}, the words in question texts on the underline are related to the knowledge concepts which on the underline with the same color. To capture the associations between question texts and knowledge concepts, we use an attention mechanism to learn the relevance between question texts and concepts. Totally, we design an attention-based LSTM to integrate question texts and knowledge concepts for diagnosing question difficulty $b$, and the architecture of attention-based LSTM is shown in Figure~\ref{fig:att-lstm}. Specifically, the sequence input to this LSTM is $x=(x_1,x_2,...,x_N)$, where $N$ is the max step of the attention-based LSTM. The hidden state $h_t$ at $t$-th step is defined as:
\begin{equation}
\begin{aligned}
&i_t=\sigma(W_{\text{xi}}x_t+W_{\text{hi}}h_{t-1}+b_i),\\
&f_t=\sigma(W_{\text{xf}}x_t+W_{\text{hf}}h_{t-1}+b_f),\\
&o_t=\sigma(W_{\text{xo}}x_t+W_{\text{ho}}h_{t-1}+b_o),\\
&c_t=f_tc_{t-1}+i_t\cdot tanh(W_{\text{xc}}x_t+W_{\text{hc}}h_{t-1}+b_c),\\
&h_t=o_ttanh(c_t),
\end{aligned}
\label{eq:lstm}
\end{equation}
where $i_*,f_*,c_*,o_*$ are the input gate, forget gate, memory cell and output gate of LSTM, and $W_*,b_*$ are parameters of the corresponding gate or cell respectively. The $t$-th input step of attention-based LSTM is defined as follow:
\begin{equation}
\setlength{\abovedisplayskip}{1pt}
\setlength{\belowdisplayskip}{1pt}
x_t=\sum_{k_i\in \mathcal{K}_q}\text{softmax}(\frac{\xi_j}{\sqrt{d_0}})k_i + w_t,\quad \xi_j = w_t^Tk_i,
\label{eq:lstmx}
\end{equation}
where $\sqrt{d_0}$ is the scaling factor~\cite{vaswani2017attention}. $\xi_j$ is the relevance between word $w_t$ and the knowledge concepts in $\mathcal{K}_q$, and $\xi_j$ is regarded as the depth and width of the concept examined by the questions, it is calculated by ``att" part shown in Figure~\ref{fig:att-lstm}. After that, an average-pooling operation is utilized to obtain parameter $b$. Also, because the theoretical range of $b$ is $[-4,4]$, we also use the $sigmoid$ function to minus $0.5$ and multiply $8$ to meet the range requirement:
\begin{equation}
\setlength{\abovedisplayskip}{1pt}
\setlength{\belowdisplayskip}{1pt}
b=8\times(\text{sigmoid}(avgragePooling(h_N)) - 0.5),
\label{eq:difficulty}
\end{equation}
where $avgragePooling$ is an operation that calculates the mean of all elements in the vector $h_N$.
\subsubsection{Prediction Module}
The last module is prediction module. As shown before, traditional IRT can use student latent trait, discrimination and difficulty of the question to predict student performance by the logistic-like formula of IRT. In DIRT framework, in order to preserve the ability of performance prediction and the interpretation of latent trait, discrimination and difficulty in traditional IRT. We input the parameters diagnosed by deep diagnosis module into the logistic-like formula of IRT to predict the student performance on the question. In this way, we can preserve the same interpretability of parameters diagnosed by deep diagnosis module as traditional IRT, and obtain the enhancing performance prediction results which are enhanced by deep learning techniques.
\begin{table}
	\centering
	\caption{The statistics of the dataset.}
	\begin{tabular}{ccc}  
		\toprule
		Statistics & Original & Pruned \\
		\midrule
		\# of history records & 65,368,739 & 5,068,039	\\
		\# of students & 1,016,235 & 81,624	\\
		\# of questions & 1,735,635 & 13,635	\\
		\# of knowledge concepts & 1,412 & 621	\\
		\# percent of text length $\le$ 30 & / & 94\%	\\
		Avg. questions per student & / & 62.09	\\
		Avg. concepts per question & / & 1.49 	\\
		\bottomrule
	\end{tabular}
	\label{tab:statistics}
	\vspace{-1em}
\end{table}
\subsubsection{DIRT Learning}
The whole parameters to be updated in DIRT mainly exit in two parts: input module and deep diagnosis module. In input module, the parameters need to be updated contain proficiency vector $\bm{\alpha}$, question embedding weights and knowledge concept dense embedding weights $\{\bf{W_Q}, \bf{W_K} \}$. In deep diagnosis module, the parameters need to be updated contain the weights of three neural networks $\{\bf{W_{DNN_a}}, \bf{W_{DNN_{\theta}}}, \bf{W_{LSTM}} \}$ which are used to learn the latent trait, discrimination and difficulty respectively. The objective function of DIRT is the negative log likelihood function. Formally, for student $i$ and question $j$, let $r_{ij}$ be the actual score, $\widetilde{r_{ij}}$ be the score predicted by DIRT. Thus the loss for student $i$ on question $j$ is defined as:
\begin{equation}
\setlength{\abovedisplayskip}{1pt}
\setlength{\belowdisplayskip}{1pt}
\mathcal{L} = r_{ij}\text{log}\widetilde{r_{ij}} + (1 - r_{ij})\text{log}(1-\widetilde{r_{ij}}),
\label{eq:loss}
\end{equation}
In this way, we can learn DIRT by directly minimizing the objective function using Adam optimization~\cite{kingma2014adam}.
\section{Experiments}
In this section, we conduct extensive experiments to demonstrate the effectiveness of DIRT framework. First, we compare DIRT with baselines on performance prediction task to validate the effectiveness of DIRT for cognitive diagnosis by exploiting question texts. Then, we compare the performance between DIRT and representative baselines to validate the strong robustness of DIRT for rare questions. In the end, we conduct a case study to visualize the strong interpretability of DIRT.
\begin{figure*}
	\centering
	\includegraphics[height=1.4in]{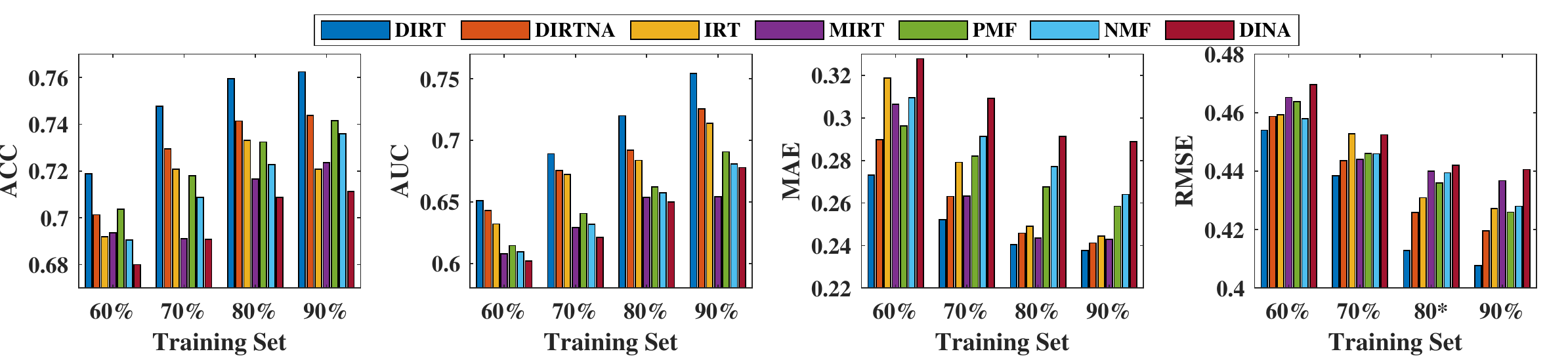}
	\caption{Overall results of student performance prediction on four metrics.}
	\label{fig:performance}
	\vspace{-1em}
\end{figure*}
\begin{figure*}
	\centering
	\includegraphics[height=1.3in]{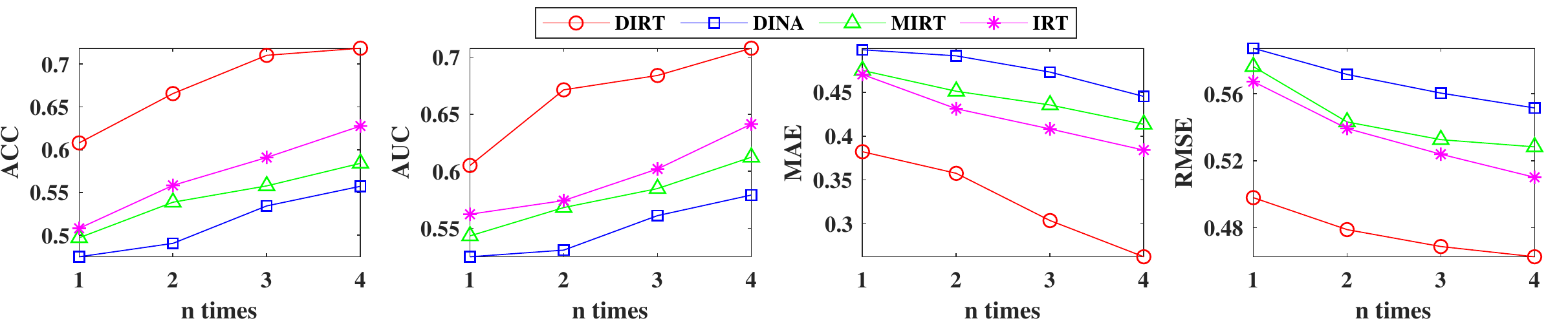}
	\caption{Results of student performance prediction on rare questions on four metrics.}
	\label{fig:cold-start}
	\vspace{-1.5em}
\end{figure*}
\subsection{Dataset Description}
The experimental dataset is composed of the mathematical data collected from a number of senior high school in China, in order to make the experimental results more reliable, we filter out the students with less than 15 questions and the questions that have no associate students. After pruning, some statistics of the dataset are shown in Table~\ref{tab:statistics}. We can observe that each student has done about 62.09 questions, and each question requires about 1.49 knowledge concepts.

\subsection{Experimental Setup}
\paragraph{Word Embedding.} The word embedding utilized in input module are trained on a large-scale mathematical question texts of the dataset. Each word of all question texts is embedded to a 50 dimension (e.g., $d_0$=50) vector by the public {\em Word2Vec} tool~\cite{mikolov2013distributed}.
\paragraph{DIRT Setting.} In DIRT, we set the maximum word sequence length $N$ of each question as 30 (padding zero when necessary) because $94\%$ of the questions are less than 30 in length shown in Table~\ref{tab:statistics}. The dimension $d_1$ of knowledge concepts dense embedding is set as 50 which is the same as $d_0$, the size of $h_t$ of attention-based LSTM is 50, respectively.

\paragraph{Training Details.} For training DIRT, we randomly initialize all parameters of DIRT framework with uniform distribution which range in $(\sqrt{-6/(n_{in}+n_{out})},\sqrt{6/(n_{in}+n_{out})})$ according to~\cite{Montavon2012Neural}, where $n_{in}$ and $n_{out}$ are the input and output size of the corresponding ones, respectively. Besides, the dropout technique~\cite{srivastava2014dropout} is also applied to prevent over fitting with probability 0.2 and the batch size is set as 32 for training.

\paragraph{Baseline Approaches.} To demonstrate the effectiveness of DIRT, we compare it with some baseline methods, the details of them are shown as follows:

\begin{itemize}
	\item {\em IRT}: IRT~\cite{embretson2000item} is a continuous cognitive diagnosis method that modeling students and questions' associated parameters by a logistic-like formula.
	\item {\em MIRT}: Different from IRT, MIRT~\cite{yao2006multidimensional} is a multidimensional cognitive diagnosis method that can model multiple knowledge proficiency of student and the parameters of question.
	\item {\em PMF}: Probabilistic matrix factorization (PMF)~\cite{mnih2008probabilistic} is a factorization method that can project students and questions into a low-rank latent factor space.
	\item {\em NMF}: Non-negative matrix factorization (NMF)~\cite{lee2001algorithms} is also factorization method, but it is non-negative which can work as a topic model.
	\item {\em DINA}: DINA~\cite{de2011generalized} is a discrete cognitive diagnosis method which is contrary to IRT, it models student concept proficiency by a binary vector.
	\item {\em DIRTNA}: DIRTNA is a variant of DIRT by only using DNN to represent question texts without attention-based LSTM.
\end{itemize}

All of the methods are implemented by PyTorch on a Linux server with four 2.0GHz Intel Xeon E5-2620 CPUs and
100G memory.

\paragraph{Evaluation Metrics.} We evaluate the performance of DIRT from two perspectives, regression perspective~\cite{willmott2005advantages}: RMSE (Root Mean Square Error), MAE (Mean Absolute Error), and classification perspective~\cite{ling2003auc}: AUC (Area Under ROC) and ACC (Prediction Accuracy).
\begin{figure*}
	\centering  
	\subfigure{
		\label{fig:proficiency}
		\includegraphics[height=1.6in]{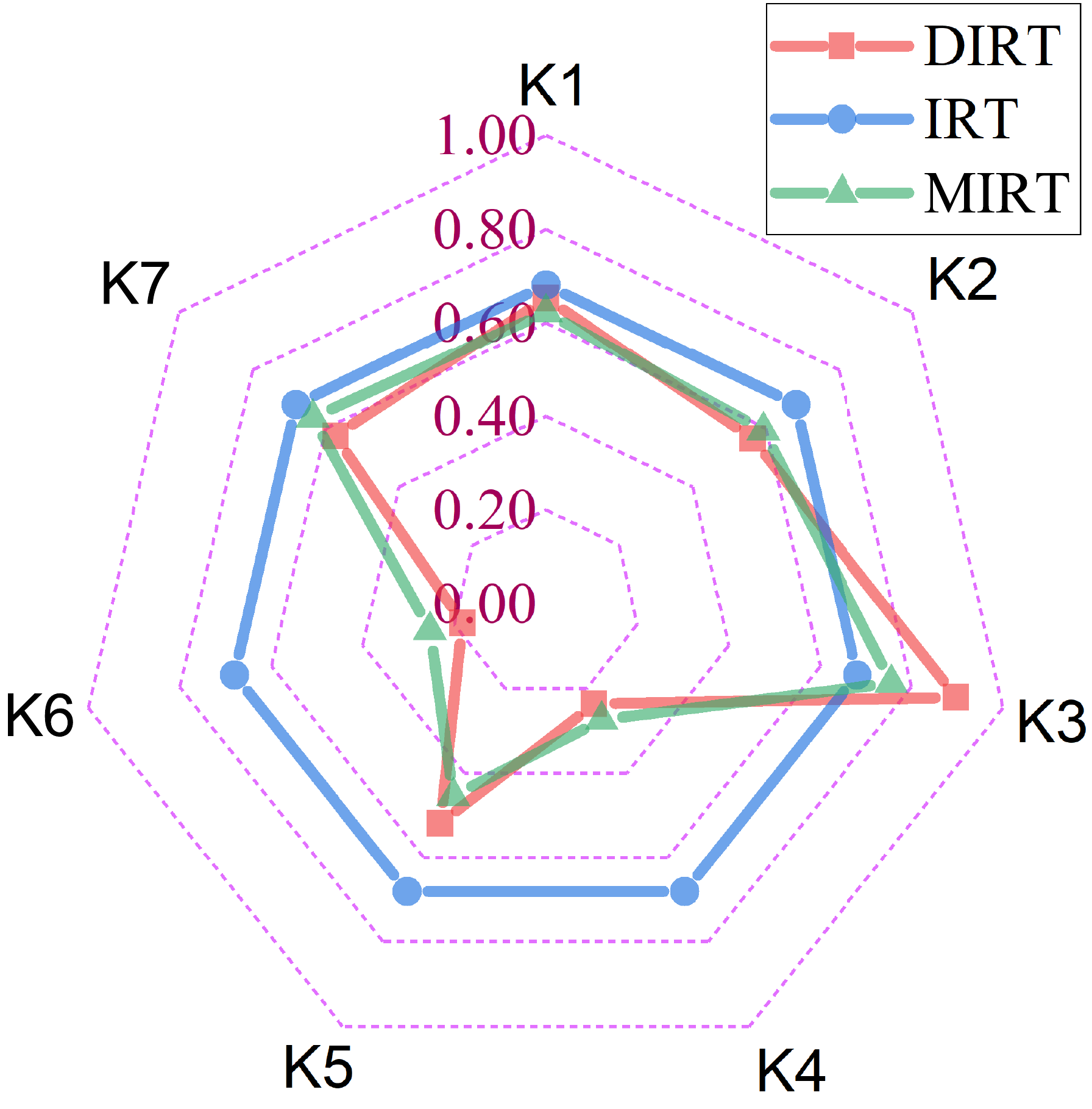}}
	\subfigure{
		\label{fig:predicting}
		\includegraphics[height=1.55in]{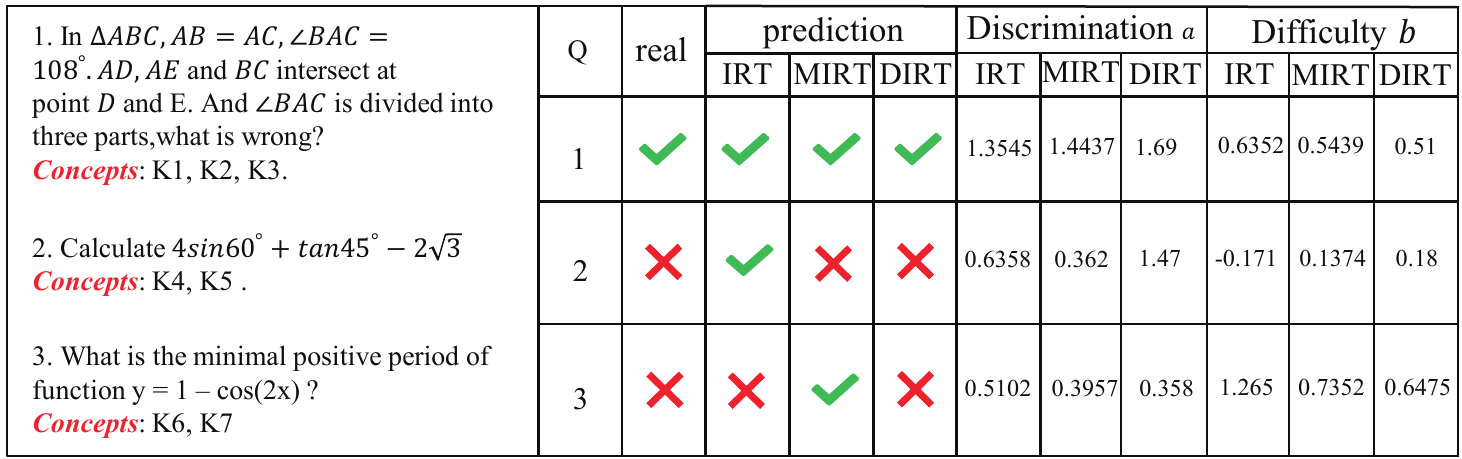}}
	\caption{Visualization of one student knowledge concept proficiency and parameters of three questions.}
	\label{fig:case-study}
	\vspace{-1em}
\end{figure*}
\subsection{Experimental Results}
\paragraph{Performance Prediction Task.} 
Here, we conduct extensive experiments on performance prediction task at different data sparsity by splitting dataset into training and testing dataset with different ratio: $60\%,70\%,80\%,90\%$. The results on all metrics are shown in Figure~\ref{fig:performance}. We can observe that compares to baselines, especially IRT, MIRT and DIRTNA, DIRT performs the best, which illustrates attention-based LSTM is effective for exploiting question texts and DIRT can provide more accurate diagnosis result to enhance traditional IRT. We can also observe that DIRT and IRT perform better than MIRT, which is mainly because MIRT is sensitive to the concept on which student has high proficiency~\cite{yao2006multidimensional}. Therefore, DIRT framework is more reliable than MIRT to the concept on which students have a high proficiency.
\paragraph{Module Robustness.} We conduct experiments to evaluate the strong robustness of DIRT for questions which only appear several times. We set three classical cognitive diagnosis models (CDMs), IRT, MIRT and DINA as representative baselines. Here, we evaluate the strong robustness of DIRT framework on the questions which only appear $1,2,3$ and $4$ times in training dataset. Figure~\ref{fig:cold-start} shows DIRT performs better than all other baselines, the results of baselines are bad, because the parameters diagnosed by them are inaccurate. This observation demonstrates that DIRT can diagnose more accurate parameters for rare questions with strong robustness. This is because deep learning techniques have great feature representation ability which is general to all question texts including the rare questions. Thus, DIRT framework has stronger robustness for rare questions.
\paragraph{Case Study.} Here, we give an example of cognitive diagnosis of student knowledge proficiency. As shown in Figure~\ref{fig:case-study}, the radar chart shows a student's concepts proficiency diagnosed by IRT, MIRT and DIRT. Because IRT only diagnose student latent trait which has the same value on all questions, we regard the latent trait in IRT as proficiency on all concepts, so the diagnosis result of IRT is a regular polygon in Figure~\ref{fig:case-study}. Therefore, DIRT can provide more accurate diagnosis results on knowledge concepts than IRT. We can also observe that DIRT predicts all three questions correctly, but IRT gets a wrong result on the second question, it is obvious that IRT gets a abnormal value of difficulty $b$ compares. Also, MIRT gets a wrong result on the third question, which is because MIRT is sensitive to concepts on which student has high proficiency~\cite{yao2006multidimensional} such as $K7$. Totally, DIRT can enhance traditional IRT with deep learning for cognitive diagnosis by exploiting question texts.

\section{Related Work}
\paragraph{Cognitive Diagnosis.} Cognitive diagnosis is wildly used in educational psychology. Many existing cognitive diagnosis models (CDM) can work well for performance prediction task. These models can be briefly divided into two types: continuous ones and discrete ones. The typical representative of the continuous models are IRT~\cite{birnbaum1968some,embretson2000item} and MIRT~\cite{yao2006multidimensional}, and they characterize each student by one  or a series of continues variables. For discrete ones, {\em deterministic inputs, noisy “and” gate model} (DINA)~\cite{haertel1984application,de2011generalized} represents each student by a binary indicator vector which indicates whether the student masters the knowledge concept or not. Recently, many researches try to improve the prediction performance, some effective models are proposed, such as {\em fuzzy cognitive diagnosis framework} (FuzzyCDM)~\cite{wu2015cognitive}.
\paragraph{Deep Learning.} Deep learning is a state-of-the-art method for many applications, e.g., image recognition~\cite{krizhevsky2012imagenet} and natural language processing~\cite{wang2016attention}. Because of the great feature representation ability\cite{mikolov2013distributed,bottou2014machine}, deep learning has been widely used to improve traditional methods. Such as DeepFM~\cite{guo2017deepfm} adopts deep neural network to improve traditional low-order FM~\cite{rendle2010factorization} model, EERNN~\cite{su2018exercise} adopts long short-term memory network to improve the performance of deep knowledge tracing (DKT)~\cite{piech2015deep}. Also, because of the strong robustness of deep learning, it also has been used to enhance the robustness of model~\cite{kim2013deep,zhang2016semantics}, such as  MANN~\cite{Liu2018FindingSE} adopts deep learning to model heterogenous data.
\section{Conclusions}
In this paper, in order to overcome the shortcomings in traditional IRT, we proposed a general DIRT framework which contained three modules: input module, deep diagnosis module and prediction module, to enhance traditional IRT with deep learning for cognitive diagnosis. Specifically, in input module, we initialized student with a proficiency vector and embedded question texts and knowledge concepts to dense vectors. In deep diagnosis module, we used deep learning to diagnosis latent trait, discrimination and difficulty in IRT by exploiting question texts.
In prediction module, we input parameters which are diagnosed by deep diagnosis module into IRT formula to predict student performance. Extensive experiments on a real-world dataset clearly validated the effectiveness and the interpretation power of DIRT. We hope our approach will lead to more relevant researches.

	\begin{spacing}{0.2}
	\bibliographystyle{named}
	\bibliography{ijcai19}
	\end{spacing}
	
\end{document}